# MODELLING AND ANALYSING CARGO SCREENING PROCESSES: A PROJECT OUTLINE

| Peer-Olaf Siebers | David Menachof | Peter Zimmerman |
| Uwe Aickelin | Galina Sherman | |

| School of Computer Science | CASS Business School | Department of War Studies |
| Nottingham University | City University London | King's College London |
| Nottingham, NG8 1BB, U.K. | London, EC1Y 8TZ, U.K. | London, WC2R 2LS, U.K. |

**ABSTRACT**

The efficiency of current cargo screening processes at sea and air ports is unknown as no benchmarks exists against which they could be measured. Some manufacturer benchmarks exist for individual sensors but we have not found any benchmarks that take a holistic view of the screening procedures assessing a combination of sensors and also taking operator variability into account. Just adding up resources and manpower used is not an effective way for assessing systems where human decision-making and operator compliance to rules play a vital role. For such systems more advanced assessment methods need to be used, taking into account that the cargo screening process is of a dynamic and stochastic nature. Our project aim is to develop a decision support tool (cargo-screening system simulator) that will map the right technology and manpower to the right commodity-threat combination in order to maximize detection rates.

In this paper we present a project outline and highlight the research challenges we have identified so far. In addition we introduce our first case study, where we investigate the cargo screening process at the ferry port in Calais.

## 1 INTRODUCTION

The primary goal of cargo screening at sea and air ports is to detect human stowaways, conventional, nuclear, chemical and radiological weapons and other potential threats. Due to the sheer volume of cargo being moved through ports between countries this is an extremely difficult task. In sea freight for example, 200 million containers are moved through 220 ports around the globe every year; this is 90% of all non bulk sea cargo (Dorndorf et al 2007).

Little is known about the efficiency of current cargo screening processes as no benchmarks exists against which they could be measured (e.g. % detected vs. % missed). Some manufacturer benchmarks are available for individual sensors, but only under laboratory conditions. It is rare to find unbiased benchmarks that have been measured in independent field tests under real world conditions. Furthermore, we have not found any benchmarks that take a holistic view of the entire screening process assessing a combination of sensors and also taking operator skills, judgment and variability into account.

Our aim is to use simulation as tool to further the understanding of cargo screening systems. However, the difficulty of simulating such processes is the fact that finding someone or something in the cargo is a rare event. For example, during the cargo screening operations at the sea port in Calais where operators search for stowaways hiding in or under lorries only approximately 0.2% of vehicles that have been checked are true positives (UK Border Agency 2009). This challenge needs to be taken into account when designing the simulation models and techniques should be considered to carefully validate the simulation. The general approach for this is to accelerate the occurrence of the rare events by using importance sampling (Heidelberger, 1995). Another research field that might provide some useful ideas in this instance is Risk Analysis, which also looks at rare events. In particular aspects of normal accident theory or high reliability theory might be useful (Sagan 1993).

In our research we attempt to identify and test innovative methods in order to advance the use of simulation for supporting decision making at the strategic and the operational level of the cargo screening process. Wilson (2005) confirms the usefulness of simulation for this and states that "Simulation modeling allows the analysis or prediction of operational effectiveness, efficiency, and detection rates (performance) of existing or proposed security systems under different configurations or operating policies before the existing systems are actually changed or a new system is built, eliminating the risk of unforeseen bottlenecks, under- or over-utilization of resources, or failure to meet specified security system requirements."

Our end goal is to have a plug and play software tool (cargo screening system simulator) that will map the right technology and manpower to the right commodity-threat combination in order to maximize the detection rate. The mapping process will be done via a multi-dimensional De-tection Rate Matrix (DRM). The development of this matrix poses another research challenge, as we might not be able to collect the required data for all the commodity-threat combinations from the real system. We want to fill the gaps in the DRM by simulating specific scenarios. Finally, we want to use the simulator to evaluate the integration of new sensor technology and to investigate how the new technology will improve detection efficiency. Simulation seems to be the ideal tool for this, as it is not possible to easily make changes to the real systems or to experiment with the real systems.

At the heart of our simulator will be a system specific DRM. To learn more about the development of such matrices we will conduct two case studies. We have chosen the case study sites based on the number of threats that are apparent on those sites. In our first case study (seaport) we look at a single threat, stowaways. In our second case study (airport) we raise the complexity of the DRM and look at multiple threats (stowaways, weapons and drugs). The research project is still at a very early stage. Therefore, we will not be able to present any results. Instead, we would like to use the workshop as a forum for a critical review of our intended research approach.

## 2  A BRIEF LITERATURE REVIEW

Simulation modeling is used to support system design and analysis. In the context of the cargo screening process, some examples (e.g. Leone and Liu 2005; Wilson 2005) have been found that use simulation modeling to evaluate key design parameters for checked baggage security screening systems in airports, in order to balance equipment cost, passenger and baggage demand, screening capacity, and security effectiveness in an attempt to meet the requirements imposed by the checked baggage screening explosive detection deadline established by the US Aviation and Transportation Security Act.

Another related subject is the enhancement of the security throughout the supply chain, i.e. achieving supply chain integrity (Closs and McGarrell 2004). Here, simulation modeling is often used to analyze the system. For example, Sekine et al (2006) use simulation and the response surface method for a trade-off analysis of port security in order to construct a set of Pareto optimal solutions.

The development of a dynamic security airport simulation is described by Weiss (2008). In contrast to the other papers mentioned so far, this simulation focuses on the human aspects in the system and employs the agent paradigm to represent the behavior of attackers and defenders. Both, attacker and defender agents are equipped with the capability to make their individual decisions after assessing the current situation and to adapt their general behavior through learning from previous experiences. This allows accounting for rapid security adaptation to shifting threads, as they might be experienced in the real world.

With regards to the DRM development a report has been found during the literature review that describes the development of such a matrix on an experimental basis (Klock 2005). The author evaluates several commercial off-the-shelf screening technologies in a controlled laboratory environment to determine their effectiveness in detecting humans in air cargo containers. The results of the study are very useful as guidance for our own DRM development, although the study has not been conducted as a field study, which we believe will have a big impact on the results, in particular the sensor performances. Furthermore, we are interested not only in the individual object but in the interaction between the different entities within the system. We will take a holistic view on the entire process to be able to consider real world dynamics and we will consider also organizational factors in our investigation.

## 3  RESEARCH CHALLENGES

As a general guidance, the decision support tool we intend to develop should help to identify the technology or combination of technologies as well as organizational processes that will provide the best possible rate of threat detection, while minimizing false positives. We have split our research questions into two categories, one focusing on the simulator development and one focusing on the DRM development.

Research questions regarding the simulator: [a] What are the boundaries of the system to be modeled? [b] How much detail do we have to model to get some meaningful output? [c] How should we model people in our system (e.g. officers or stowaways) - as simple resources or as autonomous entities? [d] How can we get a good estimate on how many stowaways, weapons or drugs are passing the borders?

In particular the last question will be quite difficult to answer. Using the Calais example again, we only know the number of stowaways failing to cross the border but not the number of those that are successful. How often do people try before they give up? There is a mathematical model that attempts to estimates the number of people that cross illegally the Mexican border (Epenshade 1995) but it fails to address the second aspect adequately.

Research questions regarding the DRM: [e] What are the most suitable classification categories to be used for the matrix we want to develop? [f] What is the best way to develop and validate a detection rate matrix for a system where the events to be modeled do not occur very frequently? [g] What is the best way to develop and validate a detection rate matrix in absence of real data or when real data is incomplete, i.e. missing data for certain technology / commodity / threat combinations? [h] Can we develop a framework to support the development of a DRM for different environments and threats?

Developing the required input distributions for modeling the detection of stowaways, weapons and other threats will be a challenge as we are dealing with rare events. We could use historic data and apply data mining as a tool to derive some patterns of occurrence but for this we would need a large amount of source data before we could draw valid conclusions. However, it is very likely that the system under observation will have changed significantly over a longer period of time (e.g. detection methods, quality of intelligence, strategies of defenders/attackers, threats), which needs to be considered in the data analysis to derive the input distributions.

## 4 DEVELOPMENT OUTLINE FOR THE DETECTION RATE MATRIX

In order to map the right technology and manpower to the right commodity thread combination we want to develop a DRM. Simulation will support this in two ways: a) by validating those parts of the matrix that are known and b) by being able to fill in gaps with estimated values derived from simulation runs. This will be a multi-dimensional construct which might have a different number of dimensions for the different application fields.

We want to start with a simple DRM and then refine it by adding more dimensions. The first DRM will have two dimensions (1) and will be developed through collecting anecdotal evidence from system insiders and where anecdotal evidence is not available by simulating specific scenarios of interest. To do the simulation exercise we first need a valid representation of the real system in form of a simulation model. As we focus on the process, traditional DES will be the simulation technique we will use for this purpose.

$$\text{rate of detection} = f \text{ (commodity \& thread combination, specific scenario)} \quad (1)$$

In order to keep the development of the first DRM simple (as this is only a proof of principle exercise) for the case study system we need for the development of the matrix, we will be looking for a system where the number of commodity / thread combinations is small. In later steps we will generalize the initial DRM (2).

$$\text{rate of detection} = f \text{ (commodity, threat, sensor)} \quad (2)$$

Sensors applicability is related to the type of commodity checked. For example, if one wants to detect stowaways in a lorry using CO2 probes which measure the level of carbon monoxide and the load consist of wood or wooden furniture which naturally exhumes carbon monoxide then the detector readings will be wrong. For this commodity the sensor is not useful and would produce many false positives, which means that in return many true negatives will stay undetected as time is wasted with manually inspecting the wrong lorries.

Next we want to add a fourth dimension to our DRM, namely cargo containment (3). This dimension will be defined by several parameters (4).

$$\text{rate of detection} = f \text{ (cargo containment, commodity, threat, sensor)} \quad (3)$$

$$\text{cargo containment} = f \text{ (type, wall thickness, wall density)} \quad (4)$$

Type is important as some of the sensors might need to have access to the interior of the containment while others might be applicable to be used from the outside. Wall thickness and density is important as many sensors have limitations regarding the penetration of the containments, depending on the containment's properties.

There are more dimensions one could add (e.g. origin, destination, shipping company, environmental conditions of test facility location etc.). Therefore, part of the research will have to deal with the question of which are the most relevant indicators of sensor efficiency?

## 5 CASE STUDY #1

For our first case study we have chosen the ferry port in Calais (France) that links Calais with Dover (UK). This site is ideal for beginning our study as the security measures in place focus only focus on detecting one threat, illegal immigrants, or clandestines, as they are called by the UK Border Force. Clandestines are people found on a lorry with the aim to get into Britain without a passport or any other papers. These can be individuals or groups. Clandestines come in hope of a better future in Britain, the lack of national identity cards and the possibility of illegal work. When clandestines do not succeed little or no publicity is printed, thereby perpetuating the false idea that clandestines are always successful. On the other hand, for every successful clandestine arriving in Britain the word goes out that the process is successful, which seems to generate even more attempts of illegal immigration (Brown 1995).

In 2003, the UK moved its border controls to Calais and operates in the harbor area to prevent people from entering the UK. There are two check points, one run by the French and one run by the UK. Various detection methods are in use at these points: heart beat detectors, passive millimeter wave sensors, CO2 probes, and body detection dogs. Between April 2007 and April 2008 the border has been passed by more than 900.000 lorries, of which in approx 0.2% some additional human freight was found. How many clandestines were missed during these checks is unknown. Although the companies supplying the detection equipment claim detection rates in the region of 100%, the experience in the field is somewhat different.

## 6 CONCLUSIONS

As we are just at the beginning of our project, we hope that on the workshop we will get some additional ideas on how to estimate the missing data and how else we can use simulation to better understand the current situation and the impact of future investment. So far, we have built a first model of the cargo screening process of our case study system. For this we have used AnyLogic 6 and its enterprise library.

In the future we plan to implement some attacker-defender scenarios, considering the evolution of defender's and attacker's strategies over time. This would require getting away from focusing on the process and moving toward focusing on the behavior of the actors. As DES usually models people as resources it is not well suited for this task. ABS is commonly used for modeling such kind of scenarios.

## ACKNOWLEDGMENTS

This project is supported by the EPSRC, grant number EP/G004234/1 and the UK Border Agency.

## REFERENCES

Brown, D.A. 1995. Human Occupancy Detection. In *Proceedings of the Institute of Electrical and Electronics Engineers 29th Annual 1995 International Carnahan Conference on Security Technology*. 166-174. Sanderstead, UK.

Closs, D.J. and E.F. McGarrell (2004). Enhancing Security Throughout the Supply Chain. Special Report to the IBM Center for the Business of Government.

Dorndorf, U., J. Herbers, E. Panascia, H.-J. Zimmermann. 2007. Ports o' Call for O.R. Problems. *OR/MS Today*. Available via <www.accessmylibrary.com/coms2/summary_0286-30867578_ITM> [accessed February 12, 2009].

Epenshade, T.J. 1995. Using INS Border Apprehension Data to Measure the Flow of Undocumented Migrants Crossing the US-Mexico Frontier. *International Migration Review* 29:545-565.

UK Border Agency. 2009. Freight Search Figures for 2007/2008, provided by the UK Border Agency (unpublished).

Heidelberger, P. 1995. Fast Simulation of Rare Events in Queuing and Reliability Models. *ACM Transactions on Modeling and Computer Simulation*. 5(1):43-85.

Klock, B.A. 2005 Examination of Possible Technologies for the Detection of Human Stowaways in Air Cargo Containers. Final Report produced for Transportation Security Administration. Atlantic City: NJ.

Leone, K. and R. Liu. 2005. The Key Design Parameters of Checked Baggage Screening Systems in Airports. *Journal of Air Transport Management* 11:69–78.

Sagan, S.D. 1993. The Limits of Safety: Organizations, Accidents, and Nuclear Weapons. Princeton, NJ: Princeton University Press.

Sekine, J., E. Campos-Náñnez, J.R. Harrald, and H. Abeledo. 2006. A Simulation-based Approach to Trade-off Analysis of Port Security. In *Proceedings of the 38th Winter Simulation Conference*, ed L. F. Perrone, B. G. Lawson, J. Liu, and F. P. Wieland, 521-528. Monterey, Ca.

Weiss, W.E. 2008. Dynamic Security: An Agent-Based Model for Airport Defense. In *Proceedings of the 2008 Winter Simulation Conference*, ed S. J. Mason, R. Hill, L. Moench, and O. Rose. 1320-1325. Pistacaway, NJ. IEEE Press.

Wilson, D.L. 2005 Use of modeling and simulation to support airport security. *IEEE Aerospace and Electronic Systems Magazine*, 208:3–8.

## AUTHOR BIOGRAPHIES

**PEER-OLAF SIEBERS** is a Research Fellow at The University of Nottingham, School of Computer Science. His email ad-dress is <pos@cs.nott.ac.uk>.

**UWE AICKELIN** is a Professor of Computer Science at The University of Nottingham, School of Computer Science, his email address is <uxa@cs.nott.ac.uk>.

**DAVID MENACHOF** is a Senior Lecturer in Transport Economics, International Logistics and Distribution at Cit University, Cass Business School. His email address is <d.menachof@city.ac.uk>

**GALINA SHERMAN** is a Management PhD student at City University, Cass Business School. Her email address is <galina.sherman.1@cass.city.ac.uk>

**PETER ZIMMERMAN** is an Emeritus Professor of Science and Security at King's College London, Department of War Studies. His email address is <peter.zimmerman@cox.net>